%% file: arxiv.tex
\documentclass{article}
\usepackage{ijcai26}

\usepackage{times}
\usepackage{soul}
\usepackage{url}
\usepackage{multirow}
\usepackage[hidelinks]{hyperref}
\usepackage[utf8]{inputenc}
\usepackage[small]{caption}
\usepackage{graphicx}
\usepackage{amsmath,amssymb,bm}
\usepackage{amsthm}
\usepackage{booktabs}
\usepackage{algorithm}
\usepackage{algorithmic}
\usepackage[switch]{lineno}

\newcommand{\std}[1]{{\scriptsize $\pm$ #1}}

\title{Beyond Static Cropping: Layer-Adaptive Visual Localization and Decoding Enhancement}

\author{
Zipeng Zhu$^1$
\and
Zhanghao Hu$^2$\and
Qinglin Zhu$^2$\and
Yuxi Hong$^3$\and
Yijun Liu$^1$\and \\
Jingyong Su$^1$\footnote{Corresponding Authors: This project is supervised by Prof. Jingyong Su, Prof. Yulan He, and Dr. Lin Gui.}\and
Yulan He$^{2*}$\And
Lin Gui$^{2*}$
\affiliations
$^1$Harbin Institute of Technology, Shenzhen\\
$^2$King's College London\\
$^3$Harbin Institute of Technology
\emails
\{220110914, 2022210917,yijunliu\}@stu.hit.edu.cn,
sujingyong@hit.edu.cn,
\{zhanghao.hu, qinglin.1.zhu, Yulan.he, lin.1.gui\}@kcl.ac.uk
}

\begin{document}

\maketitle

\input{section/abstract}

\input{section/intro}

\input{section/rel_work}

\input{section/preliminary}

\input{section/method}

\input{section/exp}

\input{section/conclusion}

\bibliographystyle{named}

\bibliography{ijcai26}

\end{document}

%% file: section/abstract.tex
\begin{abstract}
Large Vision-Language Models (LVLMs) have advanced rapidly by aligning visual patches with the text embedding space, but a fixed visual-token budget forces images to be resized to a uniform pretraining resolution, often erasing fine-grained details and causing hallucinations via over-reliance on language priors. Recent attention-guided enhancement (e.g., cropping or region-focused attention allocation) alleviates this, yet it commonly hinges on a static “magic layer” empirically chosen on simple recognition benchmarks and thus may not transfer to complex reasoning tasks.
In contrast to this static assumption, we propose a dynamic perspective on visual grounding. Through a layer-wise sensitivity analysis, we demonstrate that visual grounding is a dynamic process: while simple object recognition tasks rely on middle layers, complex visual search and reasoning tasks require visual information to be reactivated at deeper layers. Based on this observation, we introduce Visual Activation by Query (VAQ), a metric that identifies the layer whose attention map is most relevant to query-specific visual grounding by measuring attention sensitivity to the input query. Building on VAQ, we further propose \textbf{LASER} (\textbf{L}ayer-adaptive \textbf{A}ttention-guided \textbf{S}elective visual and decoding \textbf{E}nhancement for \textbf{R}easoning), a training-free inference procedure that adaptively selects task-appropriate layers for visual localization and question answering. Experiments across diverse VQA benchmarks show that LASER significantly improves VQA accuracy across tasks with varying levels of complexity.

\end{abstract}

%% file: section/intro.tex
%相似词汇汇总：
%1. 
% visually informative / visual informativeness: 主要表达信息富集程度，往往通过一张attention map/image region的attention sum（或VAQ）表示，表征一个attention map/image region所受注意力是否很高，视觉语义是否富集。

% visual focusing: LLM的注意力主要集中在特定图像区域的现象
% visual grounding: 与visual focusing相似，但是强调ground到ground-truth object/content
% visual evidence: 视觉证据，指image中对于query有帮助的证据。visual focusing/grounding两个动词ing形式表达名词意思的时候，似乎常与visual evidence意思相近/混用。
% 混用表达：visually grounded region/ visual focusing region / visually informative region

% 2. 
% relevant to query / query-relevant / query-conditioned / activated by query / modulated by query: 
% 表达情景：query-relevant visual attention, visual attention modulated by query, visual attention activated by query
% 意思：主要出现在contrastive map表述中，即我们希望区分出与query确实相关的visual attention，把这些与query相关的attention和与query无关的attention（visual attention sinks，但之前老师说避免引入sink概念）区分开。

% 3. 
% optimal map / best map selected by VAQ / selected map / optimal layer / layer selected by VAQ: 主要表达使用VAQ选出的最优层的contrastive attention map，现在许多地方表述不一致。

\section{Introduction}
\label{sec:intro}

Large vision-language models (LVLMs) have rapidly improved image-conditioned reasoning and generation by injecting visual tokens into pretrained language models~\cite{liu2023visual,team2024chameleon}. However, they remain notoriously brittle: models often produce answers which are fluent but unfaithful to image~\cite{rohrbach2018object}. 
% A key driver is the \emph{visual token bottleneck}: to align the input image with the pre-trained visual tokens, the model must resize the image to match the resolution expected by the pre-trained visual encoder. Therefore, rich scenes are compressed into a small set of patch tokens, making fine-grained evidence (small objects, text, subtle attributes, etc) easy to miss, especially in cluttered images.
A key contributor to this failure is the \emph{visual token bottleneck}. In most LVLMs, the input image must be resized to match the resolution of a fixed visual encoder, compressing rich scenes into a limited number of patch tokens. As a result, fine-grained evidence (small objects, text, subtle attributes) is easily lost, especially in cluttered scenes.
When visual evidence is weak or incomplete, models tend to fall back on language priors, leading to confident hallucinations, which is particularly concerning in high-stakes settings such as autonomous driving~\cite{zhang2023study} and medical imaging~\cite{huang2024adapting}.

\begin{figure}
    \centering
    \includegraphics[width=1\linewidth]{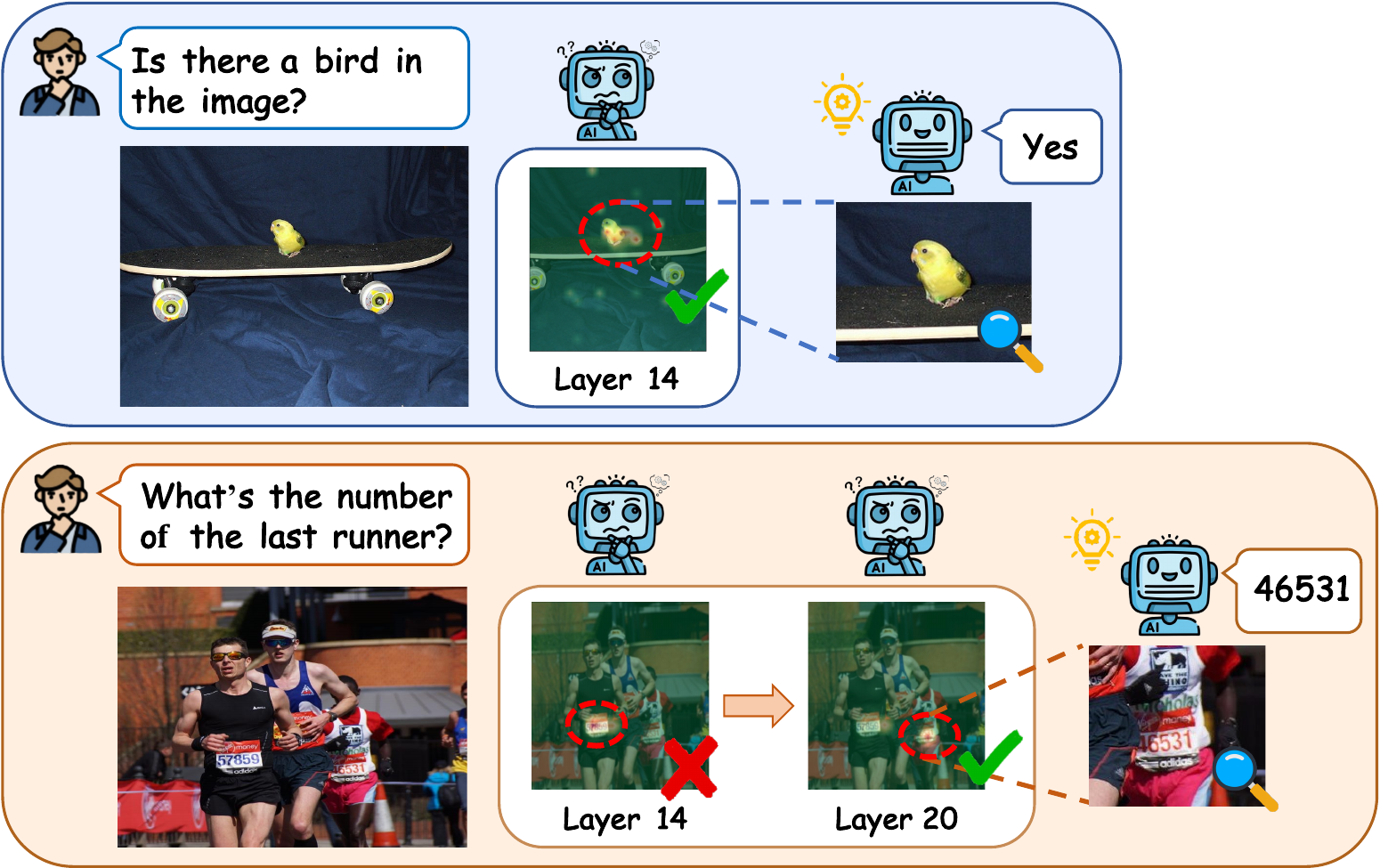}
    \caption{Visualization of LVLM attention maps across varying layers and query complexities. Red dashed circles mark the regions with the highest attention. Top: For a simple query, the model attends to the correct visual evidence effectively at a middle layer. Bottom: For a complex query requiring multi-step reasoning, the model initially focuses on visually salient but query-irrelevant regions and converges on the correct target at a deeper layer. Our work dynamically finds the correct layer to zoom in to visual evidence.}
    \label{fig:simple_complex}
\end{figure}

A popular training-free strategy to alleviate the visual token bottleneck is region refinement: localizing visually informative regions and re-feeding their crops so that the limited visual token budget is spent on the most informative content~\cite{zhang2025mllms,qi2025capturing}. 
% However, cropping is irreversible, a mis-localized crop may permanently remove supporting objects, relations, or global context. Consequently, performance can be highly sensitive to the particular attention map (and thus layer) used for localization.
% The downside is that cropping is a one-way operation. If the localization is even slightly off, the crop may remove the very evidence needed for answering the question (objects, relations, or global context), leading to brittle performance. 
However, the effectiveness of such methods depends critically on the localization signal. 
In practice, most attention-guided refinement methods extract localization cues from a fixed Transformer layer, typically in a mid-layer selected empirically on simple recognition-style tasks. This same layer is then reused unchanged for harder queries involving %in order to balance the budget and performance, most methods rely on an attention map extracted from a fixed Transformer layer; hence, crop quality depends on \emph{which layer} one uses. Nevertheless, most existing methods choose this layer heuristically, typically a fixed mid-layer selected  on simple recognition style tasks such as OCR or TextVQA, and reuses it unchanged for harder queries that require 
multi-step visual search and contextual reasoning~\cite{zhang2025mllms}. 
% to tasks requiring multi-entity reasoning and contextual inference~\cite{zhang2025mllms}.
Such practice implicitly assumes that the depth at which visual grounding emerges %``right'' grounding depth 
is task-invariant, which is hard to justify and is contrary to the layer-wise specialization observed in text-only Transformers, where layers specialize in different abstractions~\cite{jawahar-etal-2019-bert,tenney-etal-2019-bert,xin-etal-2020-deebert,zhou2020bertlosespatience}.
% This implicitly  assumes task invariant grounding depth, which is hard to justify and can fail when attention has not yet stabilised, and it also runs counter to the depth-wise evolution observed in text-only Transformers, where layers encode different abstractions and harder inputs often benefit from deeper computation~\cite{jawahar-etal-2019-bert,tenney-etal-2019-bert,xin-etal-2020-deebert,zhou2020bertlosespatience}.

% how we organize the paragrah: how to detect the visual evidence and use it for more robust VQA?
% detect:How to locate: select a map which's free of visual irrelevant attention (sinks), purely focus on visual evidence.
% robust VQA: counterfactual contrastive decoding. for evidence we localize in previous step, we enhance and mask it, contrasting them for robust reasoning.

This motivates a central question: \emph{Is visual grounding in LVLMs a static property of a single layer, or a dynamic process that depends on query complexity?} Our pilot observation suggests the latter. As shown in Figure~\ref{fig:simple_complex}, visual attention for simple object recognition queries aligns with the correct visual evidence at intermediate layers, whereas for complex, multi-step reasoning queries, visual attention initially focus on visually salient but query-irrelevant regions and only stabilizes on the correct evidence at a deeper layer.

Based on these observations, we treat visual grounding as a \emph{layer-wise, query-dependent process} rather than a static product of a single layer. %Our pilot examples suggest that attention can align with the relevant evidence earlier for simple recognition, but only stabilises in later layers for contextual reasoning (Fig.~\ref{fig:simple_complex}).
To detect evidence-bearing region dynamically and leverage them for more robust visual question answering in LVLMs, we need to address two crucial questions. 1) \textbf{How to locate}: which part of the input image provides decisive support for answering the query, and which layer of the model's latent representations most reliably captures this visual evidence for accurate localization. 2) \textbf{How to decode}: given such interpretable, layer-wise grounding signals, how to design an effective decoding strategy that fully exploits attention-guided visual focusing to improve answer accuracy. We address these questions from two perspectives: 1) \textbf{at the input level}, we design a method to identify query-relevant visual regions and adaptively refine the image to better support question answering; and 2) \textbf{at the output level}, we design a decoding method that selectively leverages the identified layers and their corresponding attention patterns for more accurate and visually grounded generation. %focus on the specific layer and corresponding attention to decode precisely. 

Our key idea is to \emph{isolate query-induced cross-modal visual attention from spurious, query-invariant artifacts} (e.g., attention sinks) via a simple counterfactual contrast. Specifically, 
we run the model on the same image \emph{with} and \emph{without} the query, and measure how each layer's visual attention distribution is \emph{modulated} by the query. From this contrast, we introduce a metric called \textbf{Visual Activation by Query (VAQ)}, which is an interpretable and model-agnostic metric that quantifies how strongly a layer's visual attention is conditioned by the query. VAQ enables query-adaptive selection of the layer for downstream visual localization and enhancement.
Despite its simplicity, VAQ effectively separates query-relevant visual attention from spurious, query-irrelevant attention patterns. 
To further exploit this grounding signal during inference, %; 2) on the input level, in order to utilize the resolute image for question answering, in the decoding stage, based on this VAQ, 
we introduce \textbf{Visual Activation of Tokens (VAT)} to identify the specific candidate answer tokens that are genuinely activated by the query-relevant visual evidence rather than solely by language patterns. VAT is then used to guide selective logits enhancement during decoding, promoting visually supported token predictions. Based on VAQ and VAT, we introduce \textbf{LASER}, a training-free method that performs query-aware visual localization and decoding enhancement, improving factual grounding and question answering of LVLMs. In summary, our main contributions are:
\begin{itemize}
    \item We propose \textbf{Contrastive Attention}, which features only query-relevant visual attention, and \textbf{Visual Activation by Query (VAQ)}, an interpretable metric that quantifies layer-wise visual activation and informativeness for a given query, enabling adaptive layer selection for visual localization beyond fixed mid-layer heuristics.
    %\item We further propose \textbf{Visual Activation by Tokens (VAT)}, a correct term of logits for more robust decoding by suppressing logits of tokens which are not activated by visual evidence found by our localization strategy.
    %\item Based on VAQ and VAT, we introduce \textbf{LASER}, a training-free method that performs query-aware visual localization and counterfactual verification, improving factual grounding and question answering of LVLMs.
    \item Based on VAQ, we introduce \textbf{LASER}, a training-free framework for query-aware visual localization and counterfactual verification, which further incorporates \textbf{Visual Activation by Tokens (VAT)} to promote token predictions supported by visual evidence and suppress unsupported token logits for more robust decoding.

    \item Extensive experiments on \textbf{Qwen-VL} and \textbf{LLaVA} show that our method improves models' visual grounding on RefCOCO and visual question answering on POPE, TextVQA and A-OKVQA benchmarks.
\end{itemize}

%% file: section/rel_work.tex
\section{Related Work}
\label{sec:related}

Many existing studies have explored the approaches of visual localization and enhancement at the input level, as well as visually informative heads and layers at the model level. We summarize the related work across these two areas as follows:

\subsection{Attention-Guided Visual Localization}
Attention-guided visual localization and enhancement is a prevailing line of training-free methods to improve LVLMs' visual grounding by exploiting models' internal attention to reallocate effective visual capacity at inference time. Representative approaches localize visually informative regions via cross-modal attention and refeed them at higher effective resolution after image cropping~\cite{zhang2025mllms}, or manipulate the attention budgets of input visual token without changing model weights, such as attention-guided warping ~\cite{attwarp2025}, visual token pruning ~\cite{blink2025}, and attention-steering that guides decoding toward visually grounded tokens ~\cite{he2025cracking,qi2025capturing}. Despite empirical gains, most visual focusing pipelines practice on the models' original attention map at a global ``best" transformer layer selected manually with additional experiments on certain datasets and then apply it broadly. This becomes brittle for queries of various difficulties and requiring different levels of visual search and reasoning. Our work challenges this fixed-layer heuristic by dynamically selecting visually informative layer which provide strongest grounding signal for visual grounding relevant to specific query, without requiring additional calibration of bounding boxes.

\subsection{Identify Visually Informative Heads and Layers}
Beyond cropping, several studies analyze where grounding signals reside inside LVLMs. ~\cite{kang2025localization_heads} identify a small subset of localization heads whose cross-modal attention maps provide strong grounding cues in frozen models. This work intuitively treats attention \emph{magnitude} and \emph{concentration} as key proxies for visual informativeness.
However, recent work reveals a phenomenon of \emph{visual attention sinks} ~\cite{kang2025visual_sink,luo2025sink}, where the tokens will receive consistently high and peaked attention \emph{regardless of the queried content}. This phenomenon suggests that static attention statistics can be confounded by content-invariant artifacts ~\cite{kang2025visual_sink}.
To mitigate this issue, ~\cite{woo2025avisc} calibrate attention during decoding by down-weighting sink-like tokens. Nevertheless, their sink identification is largely heuristic, by thresholding abnormally high attention, which may also suppress genuinely informative tokens that legitimately attract high attention.

% \begin{figure*}
%     \centering
%     \includegraphics[width=1\linewidth]{cons_attn.pdf}
%     \caption{Illustration of contrastive attention. Two QA on the same image with  and without the query. We compute the contrastive map $[\bm{a}^{v,q} - \bm{a}^{v,\varnothing}]_{+}$. Subtraction suppresses query-invariant sinks (red dashed regions), leaving high attention concentrated on query-relevant evidence}
%     \label{cons_attn}
% \end{figure*}

Overall, these studies support our view that grounding is \emph{not} statically distributed across layers and heads, while also highlighting a key challenge: a single attention map and static statistics on it can conflate spurious high attention patterns with real visual evidence related to specific query.
To address this, our VAQ explicitly tests whether visual attention is \emph{modulated by the query} via a non-query contrast, separating query-driven attention on visual evidence from disturbing attention patterns irrelevant to the query. VAT further pinpoints the candidate answer tokens activated by visual evidence.

%% file: section/preliminary.tex
\section{Preliminary}
\label{sec:preliminary}

\begin{figure}
    \centering
    \includegraphics[width=1\linewidth]{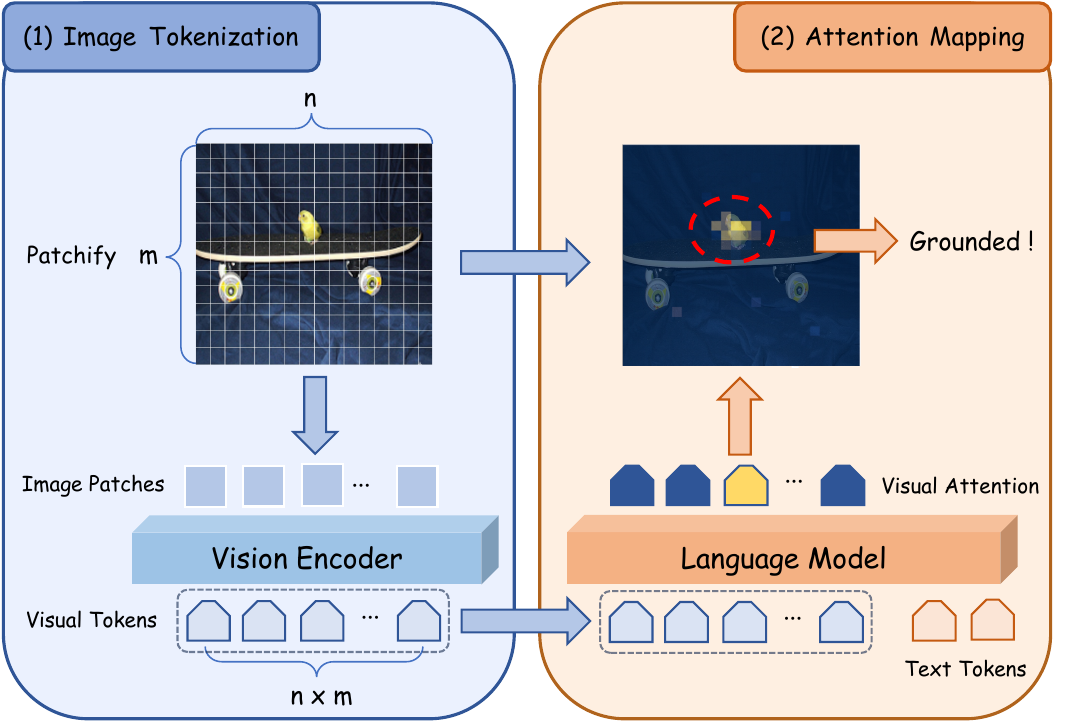}
    \caption{Illustration of the visual grounding mechanism. (1) Image Tokenization: the input image is encoded into visual tokens while explicitly preserving its $m \times n$ spatial grid structure. (2) Attention Mapping: attention weights assigned by the Language Model are projected back from the token sequence onto the 2D image plane to localize the grounded region with high attention from tokens.}
\label{fig:preliminary}
\end{figure}

In this section, we further elaborate the key mechanism of visual grounding with LVLMs' inherent attention. As illustrated in Figure~\ref{fig:preliminary}, this process consists of two stages:
\paragraph{Image Tokenization.}Following the paradigm established by ~\cite{liu2023visual}, we treat the visual input as a sequence of discrete semantic units the same as text tokens. Initially, the raw image is partitioned into a regular grid of $m \times n$ fixed-size patches. A vision encoder then projects these pixel blocks into a sequence of \emph{visual tokens}, translating high-dimensional visual signals into a latent space compatible with the Language Model. Crucially, this serialization preserves a strict spatial isomorphism: the linear sequence of visual tokens maps reversibly to the 2D coordinate system of the original image. This structural alignment serves as the physical foundation for the subsequent attention mapping.

\paragraph{Attention Mapping.}This token-patch correspondence acts as a bridge between multimodal reasoning and visual interpretability. During the generation phase, the Language Model assigns attention weights to visual tokens, effectively signaling the relevance of specific image regions to the textual query. By reshaping these linear attention scores back into the original $m \times n$ grid layout, we can project the model's internal attention focus onto the original image reversely. This process yields a spatial heatmap that localizes the visual evidence manifesting the model's ``gaze" on the grounded objects, such as the bird on the skateboard in Figure~\ref{fig:preliminary}.

These key mechanisms enable us to know \textbf{where the LVLM is looking} using attention within itself and serve as a key foundation for our cropping strategy which enhances the image input by zooming into visually grounded region.

%% file: section/method.tex
\section{Proposed Approach}

In this section, we will detail the proposed approach. The basic idea contains three steps: 1) at the model level, by using the contrast of cross-modal attention computed with and without query input to identify the important visual tokens, serving as the foundation of input visual localization and output decoding enhancement and 2) at the input level, by using the contrastive attention and identified important layer, we design a constrained visual cropping method to enhance the input image, and 3) at the output level, implemented a training-free decoding strategy in model inference. 

\subsection{Contrastive Attention and Visual Activation by Query (VAQ)}
\label{sec:val}

Driven by previous findings that many visual tokens irrelevant to queries also receive disproportionately high attention and are hard to distinguish from truly informative visual tokens with high attention, we propose a \textbf{query-conditioned contrastive attention} that isolates the part of cross-modal attention attributable to the specific query by contrasting two attention map computed with and without query.

\begin{figure*}
    \centering
    \includegraphics[width=1\linewidth]{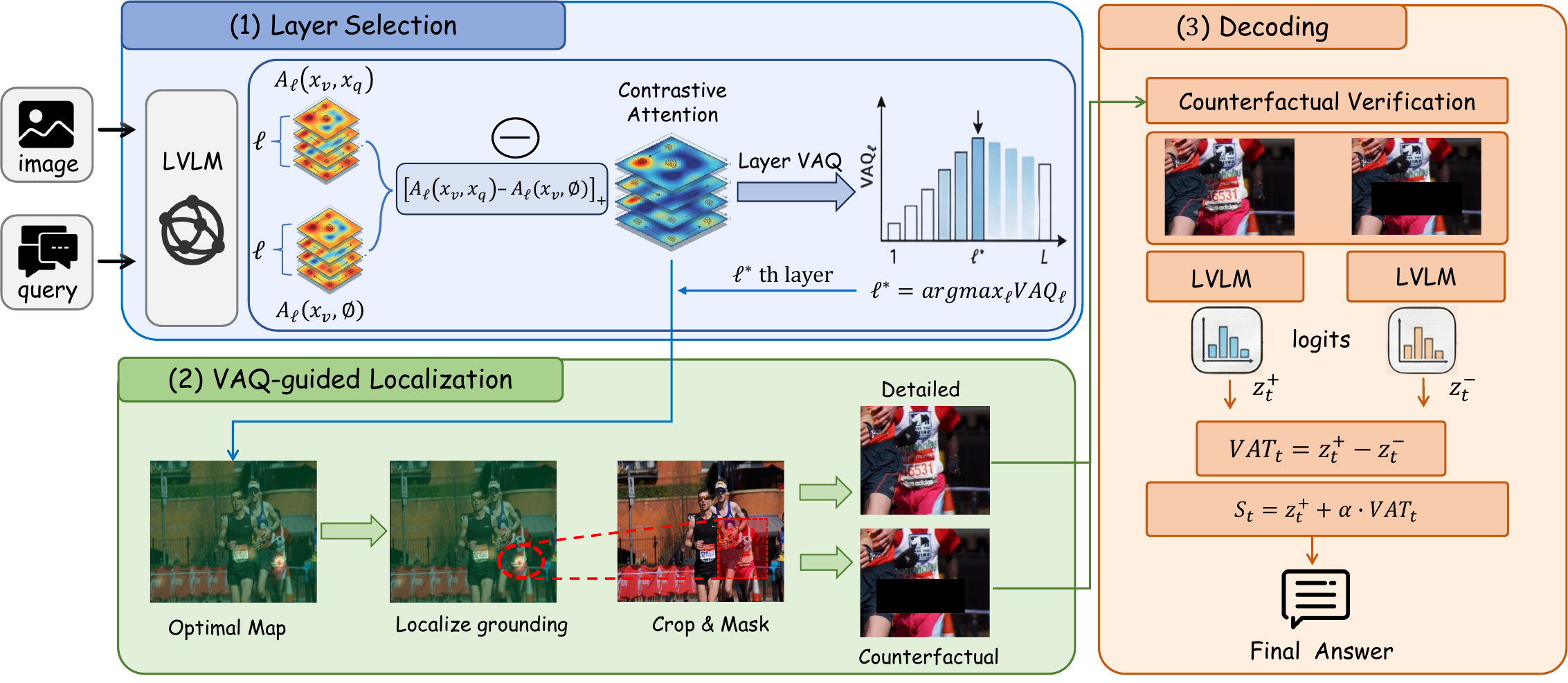}
    \caption{Overview of the LASER framework.
LASER operates in three stages to improve visual grounding and enhance image comprehension for LVLMs. (1) Layer Selection: for each layer, we compute query-conditioned contrastive attention by subtracting visual attention without the query from that with the query, yielding a layer-wise VAQ score that measures how strongly visual attention is modulated by the query; the most visually activated layer $ \ell^* $ is selected dynamically per instance. (2) VAQ-guided Localization: the contrastive attention map at $ \ell^* $ is used to localize query-relevant regions and perform constrained visual cropping, producing a focused image input that strengthens visual evidence. (3) Decoding: counterfactual verification is enabled by masking the most query-relevant visual evidence to form a negative view; logits from positive and counterfactual streams are contrasted to compute VAT and guide contrastive decoding, while easier cases decode directly from the cropped image. This design adaptively aligns visual evidence selection, localization, and decoding with query difficulty.
}
    \label{fig:framework}
\end{figure*}

% \paragraph{Contrastive Attention and Layer-wise VAQ.}
% Given the sequence of system tokens $x_{sys}$,visual tokens $ x_{vis}$ and query tokens $x_{query} $, for each layer $\ell$, head $h$, and generation step $t\in T$, the \emph{normalized} cross-modal multi-head attentions (MHA) over $P$ visual patches $\bm{a}^{v}_{\ell,h}(t)\in\mathbb{R}^{P}$  are computed twice with and without query tokens:
% \begin{align}
%     &\bm{a}^{v,q}_{\ell,h}(t) = MHA_{\ell,h}(y_t | y_{<t}, x_{sys}, x_{vis}, x_{query})\\
%     &\bm{a}^{v,\varnothing}_{\ell,h}(t) = MHA_{\ell,h}(y_t | y_{<t}, x_{sys} x_{vis}, \varnothing)
% \end{align}

% In practice, we compute ${a}^{v,\varnothing}_{\ell,h}(t)$ using the identical prompt structure, including system instructions and answer prefixes, as ${a}^{v,q}_{\ell,h}(t)$, removing only the specific query tokens. This ensures that the contrast isolates query-driven attention from generic structural priors. (See Appendix for exact templates)

\paragraph{Contrastive Attention and Layer-wise VAQ.}
Given the sequence of system tokens $x_{sys}$, visual tokens $x_{vis}$, and query tokens $x_{query}$, for each layer $\ell$, head $h$, and generation step $t\in T$, the normalized cross-modal multi-head attention weights over $P$ visual patches are computed under two conditions. We denote the visual attention weights by $v$, computed distinctively with ($q$) and without ($\varnothing$) the query tokens:
\begin{align}
    \bm{a}^{v,q}_{\ell,h}(t) &= \text{MHA}_{\ell,h}(y_t \mid y_{<t}, x_{sys}, x_{vis}, x_{query}) \\
    \bm{a}^{v,\varnothing}_{\ell,h}(t) &= \text{MHA}_{\ell,h}(y_t \mid y_{<t}, x_{sys}, x_{vis}, \varnothing)
\end{align}

In practice, we compute $\bm{a}^{v,\varnothing}_{\ell,h}(t)$ using the identical prompt structure—including system instructions and answer prefixes—as $\bm{a}^{v,q}_{\ell,h}(t)$, removing only the specific query tokens. This ensures that the contrast isolates query-driven attention from generic structural priors. (Detailed in Appendix.)

% \paragraph{Contrastive Attention and layer-wise VAQ.}
We isolate the query-induced component of each head's visual attention via a \emph{contrastive attention map}:
\begin{equation}
\mathrm{A}^{\mathrm{con}}_{\ell,h}(t)
=\Big[\bm{a}^{v,q}_{\ell,h}(t)-\bm{a}^{v,\varnothing}_{\ell,h}(t)\Big]_+
\in\mathbb{R}^{P},
\label{eq:contrastive_attn}
\end{equation}
where $[\cdot]_+$ denotes element-wise ReLU that clips negative values to zero.
Intuitively, $\mathrm{A}^{\mathrm{con}}_{\ell,h}(t)$ suppresses query-invariant sink-like patterns present in both maps and retains visual attention \emph{activated by the query} only present in $ {a}^{v,q}_{\ell,h}(t) $.

We further define the head-wise VAQ score as the magnitude of the contrastive map:
\begin{equation}
\mathrm{VAQ}_{\ell,h}(t)=\left\|\mathrm{A}^{\mathrm{con}}_{\ell,h}(t)\right\|_2.
\label{eq:vaq_head}
\end{equation}

Following prior observations that only a subset of heads are vision-relevant even within the same layer~\cite{yin2025clearsight,kang2025visual_sink},
we keep, for each layer $\ell$, the top-$K_{\text{head}}$ heads ranked by $\mathrm{VAQ}_{\ell,h}(t)$, denoted by $\mathcal{H}^{\text{top}}_\ell(t)$.
We then compute the layer-wise VAQ by averaging the top-head scores across decoding steps:
\begin{equation}
\mathrm{VAQ}_{\ell}
=\frac{1}{|T|}\sum_{t\in T}\;\frac{1}{K_{\text{head}}}\sum_{h\in\mathcal{H}^{\text{top}}_\ell(t)}\mathrm{VAQ}_{\ell,h}(t).
\label{eq:vaq_layer}
\end{equation}

In this work, we only compute the attention and VAQ at timestep $t=1$, marking the end of the prefill stage. This moment captures the model's aggregated attention focus before answering. Since the visual evidence supporting the answer to a specific visual question is stable and the model’s focus does not shift erratically during the generation of a short answer, computing attention solely at the prefill stage is sufficient. Repeating attention computation at every decoding step would only introduce unnecessary computational overhead.

\begin{figure}
    \centering
    \includegraphics[width=1\linewidth]{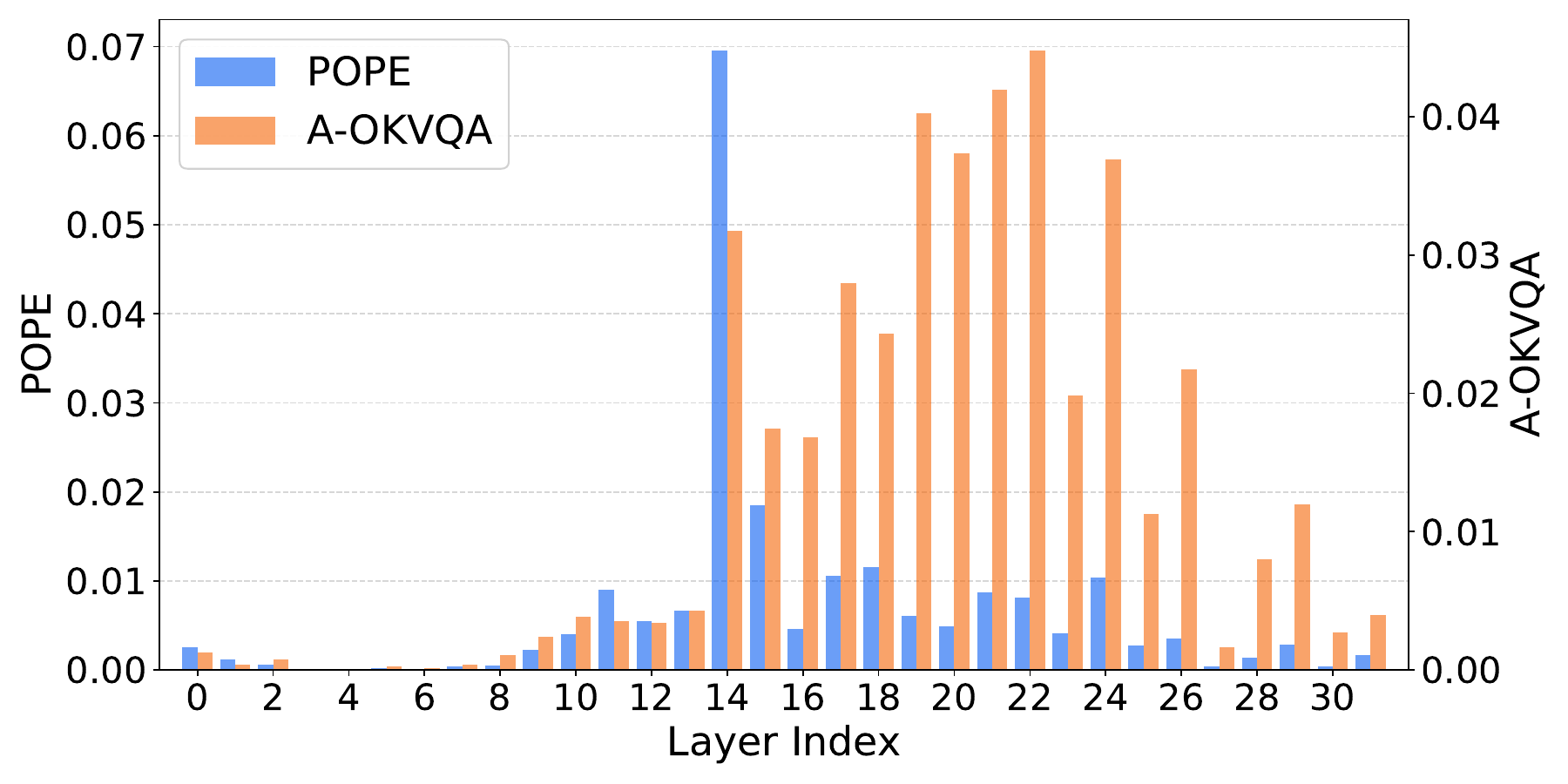}
    \caption{Layer-wise VAQ scores across queries on simple task of object existence detection (POPE benchmark) and complex task of visual reasoning (AOK-VQA benchmark). Bars show the VAQ scores at each layer averaged across different queries. Simple queries activate peaked visual attention at 14th layer while complex queries activate  much more visual attention at later layers.}
    \label{fig:vaq_bar}
\end{figure}

Layer-wise VAQ provides a profile of how strongly cross-modal attention is \emph{induced} by specific queries, as opposed to sink-like patterns that remain largely invariant to queries.
Figure~\ref{fig:vaq_bar} compares the averaged VAQ distributions on two representative benchmarks where POPE benchmark contains only simple object recognition task and A-OKVQA comprises complex reasoning tasks like spatial relation(what/where) and causal inference(why/how).
We observe a clear \emph{shift} of peak visual sensitivity across depth: POPE exhibits a sharp mid-layer spike, while A-OKVQA shows broader and generally higher VAQ in later layers.
This aligns strongly with our empirical case studies in Figure~\ref{fig:simple_complex} and indicates that the depth at which the model effectively leverages visual evidence is \emph{not} universal, but depends on the specific queries.
% Therefore, adopting a fixed, static layer for attention-guided localization is brittle across different queries.

Motivated by this, we next introduce a VAQ-guided \emph{layer selection} strategy to adaptively choose the most visually informative layer for each instance to perform enhancement.

\subsection{VAQ-Guided Image Input Enhancement via Constrained Visual Cropping}
\label{sec:convicrop}

\paragraph{Layer Selection with VAQ.}
Given $\{\mathrm{VAQ}_\ell\}_{\ell=1}^{L}$, we select the most visually activated layer $\ell^*$ with the largest VAQ, whose contrastive attention map serves as the optimal map to identify visual tokens most relevant to query and then utilize them for subsequent image enhancement.

To localize \emph{where} the visual sensitivity to the query comes from, we directly reuse the patch-wise maps $\mathrm{A}^{con}_{\ell^*,h}(t)$ in Eq.~\eqref{eq:contrastive_attn}.
Concretely, we average $\mathrm{A}^{con}_{\ell^*,h}(t)$ over the selected heads $h\in\mathcal{H}^{\text{top}}_{\ell^*}$ and over generation steps $t\in T$ to obtain a single patch-level contrastive attention map, which is a discrete distribution over image patches:

\begin{equation}
\mathrm{A}_{\ell^*}^{con}
=\frac{1}{|T|}\sum_{t\in T}\;\frac{1}{K_{\text{head}}}\sum_{h\in\mathcal{H}^{\text{top}}_{\ell^*}}\mathrm{A}^{con}_{\ell,h}(t).
\label{eq:con_attn_layer}
\end{equation}

\paragraph{Constrained visual cropping (Con-ViCrop).}
% With the optimal contrastive attention selected in Eq.~\ref{eq:con_attn_layer}, we employ visual cropping constrained by such map. 
% Let the $P$ visual patches form an $m\times m$ grid with $m=\sqrt{P}$ (e.g., $24\times24$ for $P{=}576$).
% Given the patch-level contrastive attention distribution from Sec.~\ref{sec:val}, we rank patches by mass and take the smallest prefix whose cumulative mass reaches a threshold $\tau$ (we use $\tau{=}0.9$).
% We then compute the tight axis-aligned rectangle that covers these selected patches on the grid and map it back to pixel coordinates to obtain a crop box $\mathcal{B}$ on the original image.
With the optimal contrastive attention selected in Eq.~\ref{eq:con_attn_layer}, we employ visual cropping constrained by such map. 
Following the method in Figure~\ref{fig:preliminary}, we identify the visual grounding region based on the contrastive attention map $\mathrm{A}_{\ell^*}^{con}$ and generate a crop box $\mathcal{B}_{crop}$ centered on it. 
We set the crop dimensions to half of the original image size, with a strict lower bound of $224 \times 224$. 
This minimum resolution constraint is essential because inputs which are much smaller than the CLIP visual encoder's native receptive field ($336 \times 336$) would lead to degraded token quality during the subsequent forward pass.

Finally, we crop the image to obtain $I^{+}$ and feed $(I^{+}, q)$ to the LVLM for the second-stage prediction.

% ------------------------------------------------------------

\subsection{Visual Activation of Tokens (VAT) and Contrastive Decoding}
\label{sec:token_cd}

Con-ViCrop strengthens the input by zooming into the most query-relevant region.
Following the contrastive decoding paradigm of~\cite{leng2024vcd}, which corrects token logits via a counterfactual contrast to reduce visually unfaithful predictions, we further perform a token-level counterfactual verification: we selectively obscure the \emph{most query-relevant visual evidence}, identified by the contrastive attention map at the optimal layer $ \ell^*$, and contrast the resulting logits against the logits of the uncorrupted zoomed-in image  to suppress candidate tokens unsubstantiated by visual evidence.

\paragraph{Counterfactual Input.}
Let $\ell^\star$ be the optimal layer selected by VAQ (Sec.~\ref{sec:convicrop}).
We use its contrastive attention map $A^{\mathrm{con}}_{\ell^\star}$ (Eq.~\eqref{eq:con_attn_layer}) to identify the most query-relevant visual evidence, and select the top-$K_{\text{patch}}$ patches:
\begin{equation}
\mathcal{T}
=
\mathrm{Top}\text{-}K_{\text{patch}}(A^{\mathrm{con}}_{\ell^\star}).
\label{eq:top_patches}
\end{equation}
We then mask the corresponding regions on the original image $I$ to obtain $\tilde{I}$, and apply the same crop transform $\mathcal{B}_{\text{crop}}$ to produce the counterfactual input $I^{-}=\mathrm{Zoom}(\tilde{I};\mathcal{B}_{\text{crop}})$.
This removes the visual evidence most relevant to query, yielding a strong counterfactual for token-level verification.

\paragraph{Visual Activation of Tokens (VAT).}
We run two forward passes in parallel:
\begin{align}
\quad (I^{+}, q)\ &\rightarrow\ \bm{z}^{+}_{t}\in\mathbb{R}^{|\mathcal{V}_{\text{vocab}}|},\\
\quad (I^{-}, q)\ &\rightarrow\ \bm{z}^{-}_{t}\in\mathbb{R}^{|\mathcal{V}_{\text{vocab}}|},
\end{align}
where $\bm{z}^{+}_{t}$ and $\bm{z}^{-}_{t}$ are the logits before softmax for generating the answer token at generation step $t$.

We define Visual Activation of Tokens (VAT) as the \emph{logit-level} evidence gain contributed by the visual patches containing visual evidence:
\begin{equation}
\mathrm{VAT}_{t}
=
\bm{z}^{+}_{t}-\bm{z}^{-}_{t}
\in\mathbb{R}^{|\mathcal{V}_{\text{vocab}}|}.
\label{eq:vat_logit}
\end{equation}
Intuitively, a token with a large positive $\mathrm{VAT}_{t}(v)$ is strongly supported by query-relevant visual evidence that was removed in $I^{-}$, and is more reliable in query-specific visual QA, while tokens with low  $\mathrm{VAT}_{t}(v)$ are tokens which are largely related only to language patterns and invariant to visual evidence.

We then augment the positive-stream logits with VAT:
\begin{equation}
\bm{s}_{t}
=
\bm{z}^{+}_{t}
+
\alpha\,\mathrm{VAT}_{t},
\label{eq:contrastive_score}
\end{equation}
where $\alpha>0$ controls how strongly we promote tokens that are sensitive to the removal of query-driven visual evidence.

% To reduce overhead and avoid unnecessary verification on easy queries, we apply VAT computation and contrastive decoding \emph{only when} the selected layer index satisfies
% $\ell^* > \frac{|L|}{2}$, where $|L|$ is the total number of transformer layers.
% Intuitively, a late-emerging query-conditioned visual activation indicates that the model relies on deeper, higher-level reasoning before grounding, which correlates with harder questions and a higher risk of prior-driven errors; in such cases, counterfactual verification is most beneficial.
% For $\ell^* \le \frac{|L|}{2}$, we decode with the cropped image $ I^{+}$ alone.

%% file: section/exp.tex
\section{Experiments}
We validate our approach on two representative multimodal LLM architectures and a suite of three benchmarks. These benchmarks are selected to span a spectrum of visual question answering capabilities, ranging from fundamental object existence probing to fine-grained OCR grounding, and extending to complex, open-ended reasoning.

\subsection{Experiment Setup}
\label{sec:exp_models}

\paragraph{Models.}
\textbf{LLaVA-1.5} integrates a CLIP ViT-L/14 vision encoder with an LLM via a lightweight MLP projector~\cite{liu2023visual,liu2024llava15}. 
% Specifically, after resizing different resolution of input image into a fixed resolution of $336\times336$ and partitioning it into a $24\times24$ grid of image patches, vision encoder generates a consistent number of 576 image tokens (Fig.~\ref{fig:preliminary}. 
It adopts a fixed-resolution strategy, resizing all inputs to $336\times336$ and partitioning them into a $24\times24$ grid, thereby generating a consistent sequence of 576 visual tokens (see Figure~\ref{fig:preliminary}).
In contrast, \textbf{Qwen-VL} is engineered for \emph{any-resolution} understanding~\cite{wang2024qwen2vl}. It accommodates images with varying aspect ratios by encoding them into variable-length sequences of visual tokens via a dynamic resolution mechanism.
% Improving based on LLaVA, \textbf{Qwen-VL} is engineered for \emph{any-resolution} understanding~\cite{wang2024qwen2vl}. It processes images of varying width/heigh ratios by converting them into variable length of sequences of visual tokens via dynamic resolution.
% This rigid image tokenization of both models creates a predictable one-to-one mapping between image patches and image tokens, providing an ideal testbed for our attention-guided localization and cropping. 
Crucially, despite their differences in resolution handling, both architectures preserve a deterministic \emph{spatial correspondence} between image patches and latent visual tokens. This structural alignment provides an ideal testbed for our attention-guided localization and cropping approach.

\begin{table*}[t]
\centering
\resizebox{0.87 \textwidth}{!}% optional, to resize the whole table♣
{
\begin{tabular}{llcccccccc}
\toprule
\multicolumn{2}{c}{\multirow{2}{*}{Setting}} & \multicolumn{6}{c}{POPE} & \multirow{2}{*}{TextVQA} & \multirow{2}{*}{A-OKVQA} \\
\cmidrule(lr){3-8}
\multicolumn{2}{c}{} & \multicolumn{2}{c}{Random} & \multicolumn{2}{c}{Popular} & \multicolumn{2}{c}{Adversarial} & & \\
\cmidrule(r){1-2} \cmidrule(lr){3-4} \cmidrule(lr){5-6} \cmidrule(lr){7-8} \cmidrule(lr){9-9} \cmidrule(l){10-10}
Model & Decoding & Acc. & F1 & Acc. & F1 & Acc. & F1 & Acc. & Acc. \\
\midrule
\multirow{4}{*}{LLaVA-1.5} 
 & Sample & 86.29 & 85.01 & 85.83 & 84.19 & 83.70 & 82.23 & 42.82 & 23.63 \\
 & VCD    & 87.74 & 87.08 & 86.75 & 86.26 & 85.30 & 84.50 & 44.16 & 25.54 \\
 & ViCrop& 88.52 & 87.79 & 87.73 & 86.71 & 85.80 & 84.93 & 47.04 & 23.95 \\
 & LASER(ours)   & \textbf{89.86} & \textbf{88.60} & \textbf{88.26} & \textbf{87.67} & \textbf{85.86} & \textbf{85.56} & \textbf{53.92} & \textbf{28.18} \\
\midrule
\multirow{4}{*}{Qwen-VL} 
 & Sample & 86.15 & 85.20 & 84.53 & 83.32 & 81.40 & 80.56 & 60.42 & 59.64 \\
 & VCD    & 87.28 & 86.73 & 85.83 & 85.21 & 83.15 & 82.79 & 61.18 & 61.34 \\
 & ViCrop & 86.81 & 86.45 & 85.29 & 84.96 & 82.57 & 82.28 & 62.60 & 60.12 \\
 & LASER(ours)   & \textbf{88.31} & \textbf{87.69} & \textbf{86.20} & \textbf{85.61} & \textbf{83.60} & \textbf{83.36} & \textbf{65.36} & \textbf{62.82} \\
\bottomrule
\end{tabular}
}
\caption{Performance comparison on POPE, TextVQA, and A-OKVQA benchmarks. Evaluation on A-OKVQA benchmark is run with Direct Answer setting, without providing choices.  We report Accuracy and F1 score for POPE, and Accuracy for others.}
\label{tab:results}
\end{table*}

\paragraph{Benchmarks and Datasets.} To comprehensively validate our approach, we employ a suite of datasets covering object localization and visual question answering. 
For localization, we employ \textbf{RefCOCO}, \textbf{RefCOCO+}, and \textbf{RefCOCOg} for evaluation~\cite{refcoco}. Given that these datasets provide ground-truth bounding boxes for objects described by natural language queries, we quantify the model's localization ability using the \textit{Attention Aggregation} metric. This is calculated as the sum of attention weights falling within the target object's bounding box normalized by the sum of weights across the entire attention map.
For visual question answering, we utilize three VQA benchmarks to evaluate our entire method across varying levels of granularity: \textbf{POPE}, which serves as a baseline for object hallucination robustness and factual grounding~\cite{li2023pope}; \textbf{TextVQA}, which acts as a stress test for complex visual search and isolating fine-grained visual evidence and scene text~\cite{singh2019textvqa}; and \textbf{A-OKVQA} (Direct Answer setting), which validates our depth-aware hypothesis through complex questions requiring external knowledge~\cite{schwenk2022aokvqa}.

\paragraph{Baselines.}
For localization, we compare our contrastive attention (Contrastive Att) with raw attention (Raw Att) and relative attention (Rel Att)~\cite{zhang2025mllms}. Raw Att uses model's original attention map from a pre-selected fixed layer. Rel Att, which also practices on the fixed layer, improves on Raw Att by comparing attention maps of specific and generic input. 
For visual question answering, we compare LASER against two representative training-free baselines that feature contrastive decoding and attention-guided cropping respectively.
\textbf{Visual Contrastive Decoding (VCD)} \cite{leng2024vcd} reduces language-prior hallucinations by contrasting the output logits produced from the original image and a visually noised counterpart, and then reweighting tokens during decoding based on the logit difference.
\textbf{ViCrop (Rel Att)} \cite{zhang2025mllms} is an attention-guided visual cropping method based on Relative Attention.

\paragraph{Implementation Details.}
% As introduced in Sec.~\ref{sec:val}, our VAT computation and contrastive decoding verification are applied only when the selected optimal layer satisfies $\ell* > |L|/2$. The distribution of LLaVA's optimal layer selected in different queries across three benchmarks are shown in Fig.~\ref{fig:distri_layer}, contrastive decoding is applied only for queries whose optimal layer $ \ell^* > 15 $.
For all methods cropping with static layer, we designate the 14th layer empirically according to previous work~\cite{zhang2025mllms}.
For methods that use contrastive decoding including the VCD baseline and our verification step, we set the contrast strength to $\alpha{=}1$ (Eq.\ref{eq:contrastive_score}) for a fair comparison. Temperature is set to 1 for sampling. 

\subsection{Results}
\label{sec:exp_results}

\paragraph{Visual Question Answering.}
Across three Visual Question Answering benchmarks, our method consistently outperforms baselines that either rely on cropping with attention at static layer or contrastive decoding simply with noised image. (Table~\ref{tab:results}). On POPE, even though most queries select same optimal layer $ \ell^*$ as designated 14th layer for static cropping method, we still surpass ViCrop by a clear margin, indicating our contrastive map provide more reliable grounding signals than previous methods. On TextVQA, questions often require multiple steps of localization and fine-grained text reading, the optimal grounding layer varies substantially (Figure~\ref{fig:distri_layer}). Our VAQ-based dynamic layer selection remains robust and outperforms ViCrop. On the most challenging A-OKVQA, where mis-cropping can cause irreversible information loss, we still achieve a noticeable improvement, and the higher frequency of VAT/contrastive decoding indicates  decoding  further helps suppress language-prior answers when grounding is difficult. 
Notably, improvements of our method and ViCrop on LLaVA are substantially larger than on Qwen-VL, which is expected: LLaVA resizes high-resolution images to a fixed resolution, so cropping effectively zooms in and reallocates limited pixels to regions relevant to queries, whereas Qwen-VL’s any-resolution encoding preserves fine details across the whole image. Still, cropping remains beneficial for any-resolution models in practice, since encoding the full extremely high resolution image and computing attention over all visual tokens incurs heavy GPU memory and runtime overhead; cropping reduces the number of visual tokens that must be encoded, making inference more efficient without sacrificing performance due to limited resolution.

\begin{table}[htbp]
    \centering
    \small % 整体字号 small
    \setlength{\tabcolsep}{4pt} 
    
    \begin{tabular}{llccc}
        \toprule
        \multicolumn{2}{c}{\textbf{Setting}} & \multicolumn{3}{c}{\textbf{Benchmarks}} \\
        \cmidrule(r){1-2} \cmidrule(l){3-5}
        Model & Method & RefCOCO & RefCOCO+ & RefCOCOg \\
        \midrule
        \multirow{3}{*}{LLaVA} 
            & Raw Att    & 28.27 \std{1.07} & 24.99 \std{0.98} & 22.86 \std{0.99} \\
            & Rel Att    & 30.96 \std{1.11} & 27.29 \std{0.98} & 25.34 \std{0.96} \\
            & Contra Att & \textbf{41.77} \std{1.43} & \textbf{38.44} \std{1.44} & \textbf{31.92} \std{1.36} \\
        \midrule
        \multirow{3}{*}{Qwen-VL} 
            & Raw Att    & 25.40 \std{1.05} & 24.53 \std{1.02} & 23.06 \std{1.15} \\
            & Rel Att    & 31.34 \std{1.17} & 30.27 \std{1.14} & 28.92 \std{1.20}  \\
            & Contra Att & \textbf{34.36} \std{1.39} & \textbf{34.81} \std{1.43} & \textbf{30.56} \std{1.46}  \\
        \bottomrule
    \end{tabular}
    \caption{Localization performance comparison on RefCOCO(+/g). We report the mean and standard error of Attention Aggregation Ratio (\%) over all image-query pairs in the dataset. Raw Att represents the original raw attention map of model at the fixed layer. Rel Att represents relative attention at the same fixed layer. Contra Att represents our contrastive attention with dynamic layer selection strategy. }
    \label{tab:refcoco}
\end{table}

\paragraph{Localization.}
Attention aggregation on reference visual content is a crucial sign for reference object localization as shown in Sec.~\ref{sec:preliminary}. Therefore, we use Attention Aggregation, the visual attention sum ratio aggregated inside the annotated grounding box, to measure the magnitude of visual attention that correctly focus on the reference object. As shown in Table~\ref{tab:refcoco}, Attention Aggregation ratio of our method is higher than that of raw attention and relative attention, both with static layer. Our dynamic layer selection strategy selects only 32.55\% of the designated static layer but still show higher Attention Aggregation, demonstrating that the optimal attention map to localize are indeed from different layers and VAQ selection are able to select it efficiently.

\begin{table}[h]
\centering
\begin{tabular}{l|cc|cc} % 5 列
\toprule
& \multicolumn{2}{c|}{\textbf{LLaVA-1.5}} & \multicolumn{2}{c}{\textbf{Qwen-VL}} \\
\cmidrule(lr){2-3} \cmidrule(lr){4-5}
& Acc. & Time Cost & Acc. & Time Cost \\
\midrule
Sample     & 42.82 & 655.83 & 60.42 & 291.81 \\
ViCrop     & 47.04  & 826.73 & 62.60& 369.50 \\
\midrule
LASER     & 53.92 & 874.41 & 65.36 & 417.47 \\
{\bf w/o} VAT      & 49.34 & 850.61 & 64.28 & 410.93 \\
{\bf w/o} VAQ/VAT  & 47.28 & 820.96 & 62.54 & 378.15 \\
\bottomrule
\end{tabular}
\caption{Ablation study on TextVQA. We compare the Accuracy and Time Cost (ms) across two different base models (LLaVA-1.5 and Qwen-VL). w/o VAT disables VAT decoding based on LASER. w/o VAQ/VAT further disables dynamic layer selection with VAQ, but still uses contrastive attention at the fixed 14th layer.}
\label{tab:ablation}
\end{table}

\subsection{Further Analysis}

\begin{figure}
    \centering
    \includegraphics[width=1\linewidth]{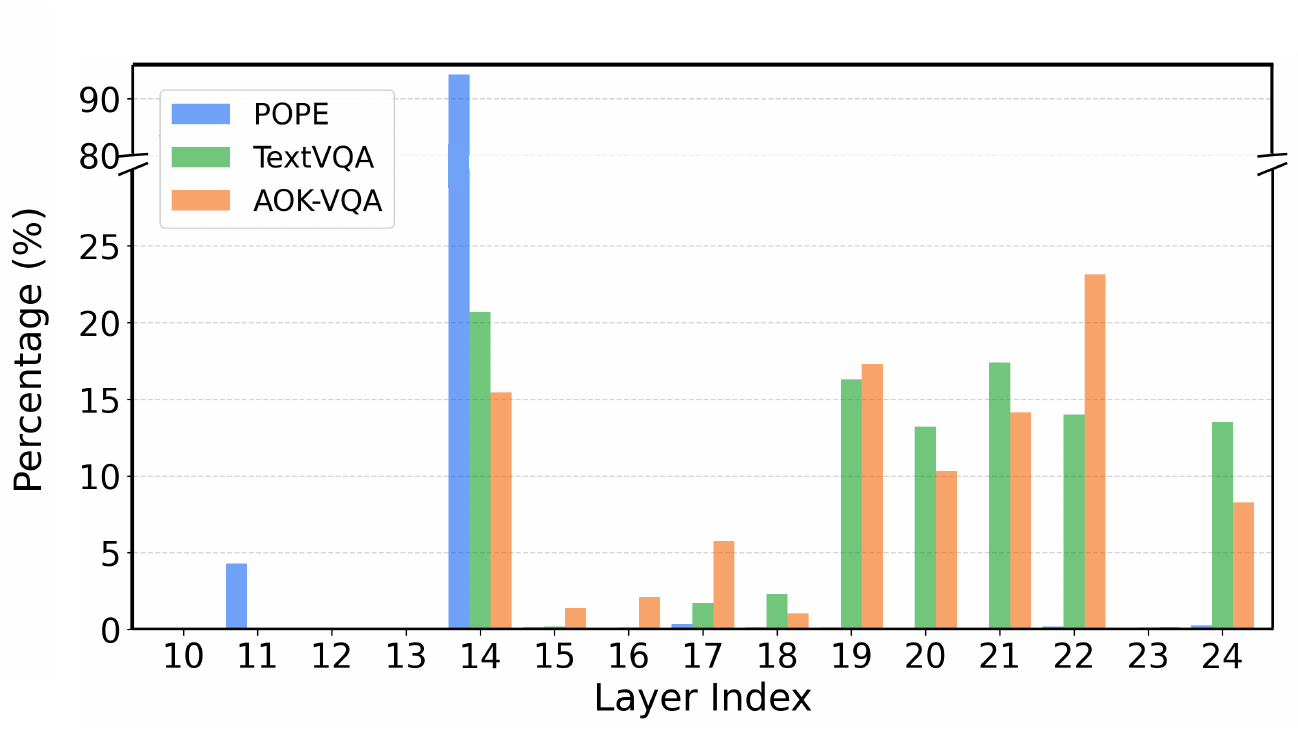}
    \caption{The distribution of the most visually activated layers selected by VAQ across different queries in three benchmarks. The 14th layer is selected overwhelmingly in POPE while later layers are more frequently selected in other two benchmarks.}
    \label{fig:distri_layer}
\end{figure}

% 这里还要改 TODO:
\paragraph{Dynamic Selection of Layer.} 
Figure~\ref{fig:distri_layer} shows the distribution of layers dynamically selected by VAQ in three benchmarks. The same as our previous case studies (Figure~\ref{fig:simple_complex}), for direct object existence detection such as POPE, the LVLMs localize the objects in mid-early layer overwhelmingly. However, when the query requires multiple steps of visual search or complex contextual reasoning such as in TextVQA and A-OKVQA, LVLMs attend to the correct visual evidence related to queries more at later layers. This also explains why our methods outperform ViCrop more on harder tasks such as TextVQA and A-OKVQA than simple task like POPE. 

\paragraph{Ablation Study.} 
% As shown in Fig.~\ref{fig:distri_layer}, the optimal layers $ \ell*$ selected with VAQ for cropping  vary among queries in TextVQA and A-OKVQA, and queries whose $ \ell*>|L|/2$ trigger VAT computation and contrastive decoding. 
To validate the efficacy of both VAQ-based dynamic layer selection and VAT-based contrastive decoding, we conduct ablation study on LLaVA with TextVQA as shown in Table~\ref{tab:ablation}. We compare w/o VAQ/VAT and w/o VAT settings which respectively stand for cropping with a fixed layer 14 and cropping with a layer selected with VAQ. While cropping with fixed layer already yields satisfactory result, VAQ and dynamic layer selection further improve the performance. This aligns with the optimal layer distribution shown in Figure~\ref{fig:distri_layer}, that 14th layer is the most selected layer with VAQ in TextVQA for optimal cropping but there are still considerable proportion of queries that require deeper reasoning and are visually grounded in later layers. Additional VAT and decoding strategy further improve the accuracy by suppressing answer tokens unsupported by visual evidence.

\paragraph{Time Cost.} The time costs of baselines and our method on TextVQA benchmark are shown in Table~\ref{tab:ablation}. Tests of time cost are conducted on two NVIDIA GeForce RTX 4090 and support concurrent computation using multiple GPU threads. While our method requires two additional attention passes and a counterfactual decoding branch, they are both parallelizable and computed concurrently. Note that our computation overhead {\bf will not scale} with the number of generated tokens, as two attention maps for contrastive attention are only computed with input tokens during the prefill stage without recomputing with generated tokens.
On TextVQA where numbers of generated tokens for answer are often from 2 to 4, the full setting of LASER incurs additional 33.33\% time cost on LLaVA-1.5 and 43.06\% on Qwen-VL, given twice the GPU memory for concurrency.

%% file: section/conclusion.tex
\section{Conclusion}

This work challenges the prevailing static-layer assumption in attention-guided visual enhancement for Large Vision-Language Models by revealing that visual grounding is inherently dynamic across layers. Through comprehensive layer-wise sensitivity analysis, we show that while mid-level layers are sufficient for simple object localization tasks, deeper layers must re-engage visual information to support multi-step grounding and complex reasoning. Building on this insight, we introduce Visual Activation by Query (VAQ) to identify query-relevant visual grounding layers, Visual Activation of Token (VAT) to identify candidate generated tokens which are reasonably activated by visual evidence and propose LASER, a training-free, layer-adaptive inference framework for selective visual localization and decoding enhancement. Extensive experiments across diverse VQA benchmarks demonstrate that LASER consistently improves performance across varying reasoning complexities, highlighting the importance of dynamic, query-aware visual grounding in LVLMs.

\section*{Acknowledgment}

This work was supported by King’s Computational Research, Engineering, and Technology Environment (CREATE) and was supported in part by the UK Engineering and Physical Sciences Research Council (EPSRC) through a Turing AI Fellowship (grant no. EP/V020579/1, EP/V020579/2), National Natural Science Foundation of China (No. 62576120), and PhD studentships from the Chinese Scholarship Council funds Zhanghao Hu, and Qinglin Zhu.